# An Empirical Evaluation of Quantum Kernel Stability in Network Intrusion Detection

1st Sreejonty Chakraborty
*Department of Computer Science and Engineering*
*Bangladesh University of Business and Technology*
Dhaka, Bangladesh
sreejonty8012@gmail.com

2nd Adrita Rahman Tory
*Department of Computer Science and Engineering*
*Bangladesh University of Business and Technology*
Dhaka, Bangladesh
adritatory@bubt.edu.bd

3rd Muhammad Aminur Rahaman
*Department of Data Science and Engineering*
*Bangladesh University of Business and Technology*
Dhaka, Bangladesh
aminur@bubt.edu.bd

4th Khondokar Fida Hasan*
*Cybersecurity Discipline, School of Professional Studies*
*University of New South Wales (UNSW)*
Canberra, Australia
fida.hasan@unsw.edu.au

***Abstract*—Quantum kernel methods have been proposed for network intrusion detection, yet their performance limitations are often attributed to noise in current quantum hardware. This study evaluates that assumption through a controlled analysis of Pegasos-QSVC on the NSL-KDD and ToN-IoT datasets. We perform 56 quantum experiments using one exact noiseless statevector backend and three noise-simulated superconducting backends while keeping preprocessing, PCA features, data splits, and classical baselines identical. XGBoost achieves accuracies of 98.94% on NSL-KDD and 99.41% on ToN-IoT, compared with 91.80% and 82.98% for Pegasos-QSVC. The quantum model also exhibits substantially greater seed-dependent variability. Eight runs collapse into degenerate classifiers, including four recall-collapse cases and four trivial-positive classifiers. Notably, two failures occur on the noiseless backend, demonstrating that hardware noise is not a necessary condition for classifier collapse. Kernel Gram matrix analysis further reveals low effective rank and severe ill-conditioning even in the absence of noise, indicating intrinsic fragility in the learned quantum representation. These results show that average accuracy alone can conceal operationally unacceptable behavior and highlight the importance of reporting worst-case performance, seed-level variability, class-specific metrics, and explicit degeneracy tests in quantum intrusion detection studies.**

## I. Introduction

Intrusion detection systems are expected to operate at a consistent, predictable level of performance, not just a high average one. NIST identifies stable, tunable detection behavior as a core requirement for intrusion detection and prevention systems, since operators calibrate alert thresholds and staffing around an expected detection rate [1]. Classical machine learning has been applied to this problem for two decades, and survey work shows these methods now deliver largely consistent results across a range of network conditions [2]. Hybrid classical pipelines combining feature selection with gradient-boosted models have reported accuracy above 99% on standard intrusion detection benchmarks [3].

Quantum machine learning has recently been applied to the same task. Quantum kernel methods, introduced by Havlicek et al. as a way to map classical data into richer feature spaces using quantum circuits [4], have motivated a growing number of intrusion detection studies. Cirillo and Esposito trained a Pegasos quantum support vector classifier [5], a variational quantum classifier, and a hybrid quantum classical network on ToN-IoT and NSL-KDD, and reported that Pegasos-QSVC with a ZFeatureMap reached 93.05% accuracy averaged across four random seeds [6]. Gouveia and Correia reported an earlier quantum assisted detector whose accuracy varied considerably depending on the dataset used [7]. This growing body of work is generally optimistic about the near-term potential of quantum methods for this task.

Most of these studies attribute shortfalls to hardware noise. When a quantum model underperforms a classical baseline, the gap is typically explained by gate error, decoherence, and finite measurement shots inherent to current NISQ-era devices [8], with the implied conclusion that better hardware will close the gap [6]. This explanation can be tested directly. Running the same model on a noiseless simulator removes gate error and decoherence entirely. If performance recovers, hardware noise was the cause. If it does not, the explanation does not hold, and better qubits will not fix it.

This work makes three primary contributions:

- We conduct a matched evaluation of Pegasos-QSVC across noiseless and noise-simulated backends and of classical baselines (XGBoost and SVM) on NSL-KDD and ToN-IoT and found that the role of hardware noise is not primary in performance instability.
- We identify and characterize two distinct forms of seed-dependent classifier degeneration: recall collapse and trivial-positive classification, occurring in 8/56 runs in-

Corresponding author, Email: fida.hasan@unsw.edu.au

cluding 2 on the noiseless backend. These demonstrate why average accuracy alone can conceal operationally unacceptable behaviour in security-critical applications.

- We analyse the kernel Gram matrices to investigate the observed instability and find the noiseless kernels exhibit low effective rank and severe ill-conditioning on both datasets, providing evidence that the instability may originate partly from the representation itself. This motivates the use of worst-run performance, seed-level variability, class-specific metrics, and explicit degeneracy tests when evaluating quantum intrusion detection models.

The remainder of the paper proceeds from evidence to mechanism. Section 2 situates this work relative to the Cirillo and Esposito benchmark [6] and the quantum kernel concentration literature [9], [10]. Section 3 describes the pipeline, the backends, and the criteria used to identify failure across all 56 runs. Section 4 reports results in the order the evidence requires: accuracy, then the collapse events, then the kernel geometry that explains them. Section 5 argues that averaged accuracy alone is an insufficient reporting standard for a security critical classifier, and Section 6 states what this study did not test.

## II. Background and Related Work

A support vector machine separates classes by computing similarity between data points through a kernel function, most commonly an inner product in the input space or a fixed nonlinear transformation of it. A quantum kernel classifier replaces that inner product with a fidelity measured between two quantum states. Each classical feature vector is encoded into a parameterized circuit, and the kernel entry for two data points is the probability that their encoded states overlap [4]. The motivation for this substitution is that the encoding lives in a Hilbert space of dimension $2^n$, exponentially larger than the seven-dimensional input used here, and a richer space could, in principle, separate classes that a linear kernel cannot [6]. We use the Pegasos formulation throughout, a stochastic subgradient solver that optimizes the support vector objective in the primal and scales independently of training set size [5]. Pegasos-QSVC is also the model at the center of the strongest existing benchmark in this literature. Cirillo and Esposito evaluate Pegasos-QSVC, a variational quantum classifier, and a hybrid quantum-classical network on ToN-IoT and NSL-KDD across the FakeLagos, FakeNairobi, and FakePerth noise models, identifying ZFeatureMap as the most noise-robust encoding for this class of backend [6]. Gouveia and Correia report an earlier quantum-assisted detector whose accuracy varies considerably across datasets [7]. Their findings are consistent with the sensitivity to configuration that later, larger studies quantified. However, the Cirillo and Esposito benchmark averages results over only four random seeds, reflecting the common reporting practice in this literature. We argue that this averaging can conceal important failure modes.

Existing theory suggests that hardware noise is not the primary limitation. Thanasilp *et al.* show that fidelity quantum kernels suffer exponential concentration, a phenomenon formally related to barren plateaus in variational circuits [10], [11]. As the encoding becomes more expressive, kernel values converge toward a constant, reducing the effective information available for classification. Kübler *et al.* further show that, without engineered inductive bias, fidelity-based quantum kernels cannot learn beyond the constant function, even on a perfect noiseless quantum computer [9].

Recent empirical evidence increasingly supports these theoretical predictions. Bellante *et al.* conclude that quantum principal component analysis and clustering are unlikely to outperform classical approaches for intrusion detection under realistic assumptions [12]. Bowles, Ahmed, and Schuld benchmark twelve quantum models against classical baselines on 160 datasets and find that classical methods consistently outperform quantum alternatives, while only about 4% of prior studies report similar conclusions [13]. Most existing work attributes the performance gap to hardware limitations rather than the quantum models themselves.

TABLE I
Summary of Related Work

| Study | Dataset | Noise | Key Finding |
|---|---|---|---|
| Cirillo & Esposito [6] | ToN-IoT, NSL-KDD | 3 IBM models | Pegasos-QSVC, VQC, hybrid QNN. ZFeatureMap most robust; 94.60% acc. (4 seeds). |
| Gouveia & Correia [7] | Custom | Not reported | Quantum-assisted detector; accuracy varies with dataset. |
| Thanasilp et al. [10] | N/A | N/A | Fidelity kernels (theory); exponential concentration ≈ barren plateaus. |
| Kübler et al. [9] | N/A | N/A | Quantum kernels (theory); no bias → kernel learns constant function. |
| Bellante et al. [12] | Intrusion detection | Realistic | Quantum PCA and clustering unlikely to outperform classical methods. |
| Bowles, Ahmed & Schuld [13] | 160 datasets | Not focus | 12 quantum models; classical wins, only 4% of literature agrees. |
| **This work** | NSL-KDD, ToN-IoT | 4 backends incl. noiseless | Pegasos-QSVC; collapse without noise and Gram matrix rank collapse. |

Table I summarizes these complementary lines of evidence. Theory predicts that fidelity-based quantum kernels can fail even without hardware noise, while empirical studies show classical methods outperforming quantum approaches on average. However, no previous study has examined seed-level classifier collapse on cybersecurity datasets or demonstrated that such failures occur under noiseless simulation. This work addresses that gap.

## III. Methodology

Figure 1 summarizes the complete experimental pipeline, from dataset preparation to kernel diagnostics and classifier evaluation. Every experimental arm shares the same preprocessing pipeline, train–test split, feature representation, and evaluation protocol to ensure that observed performance differences arise solely from the learning algorithm and execution backend.

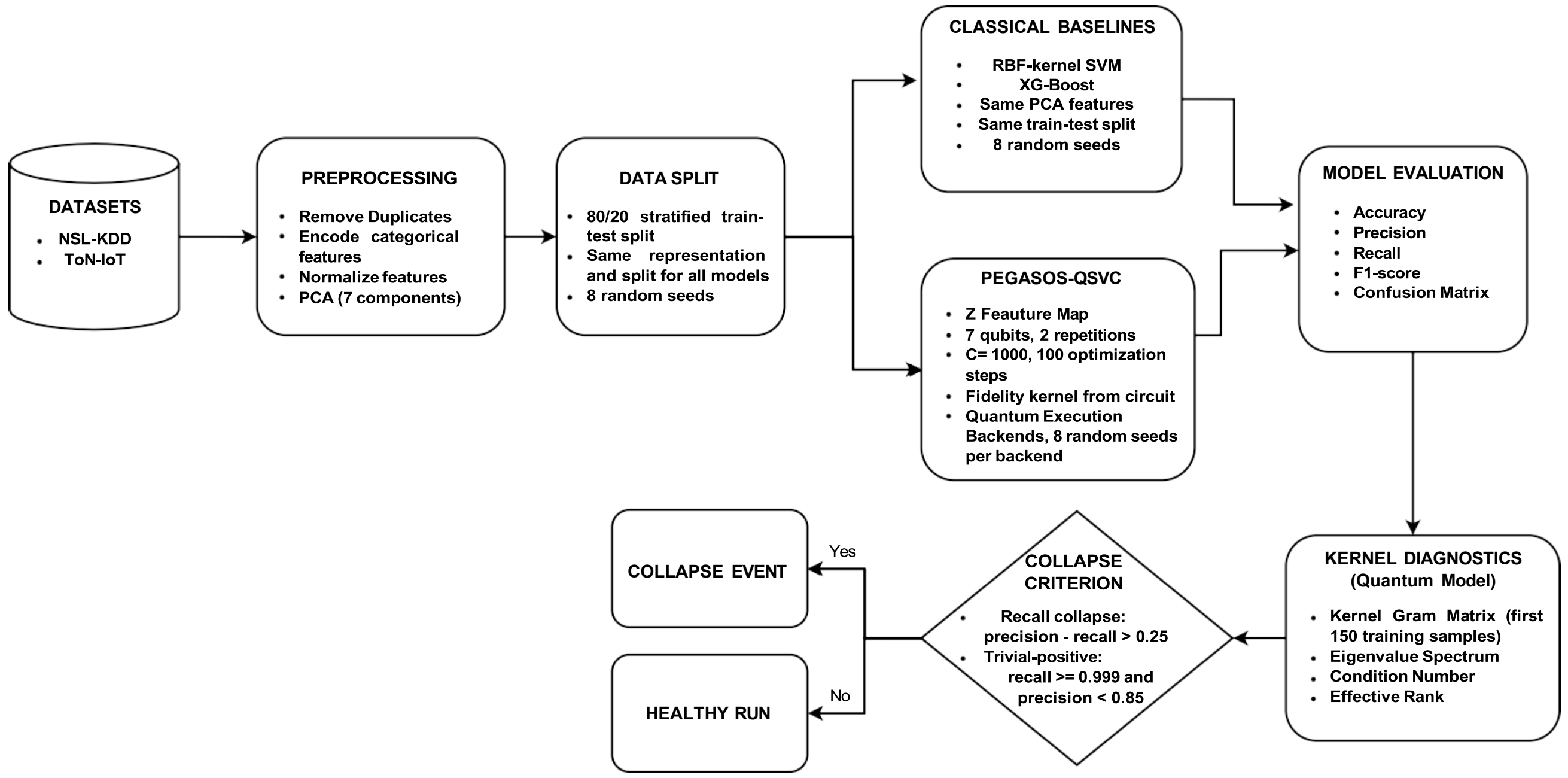

Fig. 1. Preprocessing and evaluation pipeline shared by every arm of the experiment.

**Datasets.** NSL-KDD is the corrected successor to KDD Cup 99, designed to remove the redundant records that inflated earlier evaluations [14]. We use the KDDTrain+ split, convert the attack category into a binary label, and retain all 41 original feature columns. ToN-IoT is a telemetry and network traffic dataset collected from a medium-scale IoT testbed at UNSW Canberra [15]. We note that dataset choice can materially affect how representative an evaluation is of real-world, industry-specific threats [16], our conclusions here are scoped to the attack types present in these two benchmarks. From the network traffic table (211,043 records, 44 attributes), we remove the multiclass attack-type column because it directly encodes the target label, together with seven identifier and free-text attributes (e.g., source IP, destination IP, DNS query, and HTTP URI), leaving 35 input features.

**Preprocessing.** The preprocessing pipeline is identical for every model. Duplicate records are removed, categorical attributes are encoded, numerical features are normalized, and Principal Component Analysis (PCA) reduces the feature space to seven components. An 80/20 stratified train–test split produces 3,779 test flows for NSL-KDD (46.55% attack prevalence) and 6,332 test flows for ToN-IoT (76.31% attack prevalence). The same PCA representation and data split are supplied to both quantum and classical classifiers so that any performance differences are attributable to the classifiers rather than feature engineering.

**Quantum model and execution backends.** Pegasos-QSVC employs a seven-qubit `ZFeatureMap` with two repetitions, regularization constant $C = 1000$, and 100 optimization steps, matching the configuration reported by Cirillo and Esposito as the strongest performer for this dataset family [6]. The noiseless configuration computes kernel fidelities exactly using a statevector simulator without sampling. Three additional experiments employ Qiskit Aer noise models derived from the FakeLagosV2, FakeNairobiV2, and FakePerth calibration snapshots using 1,024 measurement shots. Throughout the paper, these are referred to as Lagos, Nairobi, and Perth, respectively. Comparing the noiseless simulator with the noisy backends isolates the effect of hardware noise while keeping every other experimental variable fixed.

**Baselines, seeds, and collapse criteria.** Classical baselines include an RBF-kernel Support Vector Machine and XGBoost, each evaluated using the same eight random seeds as Pegasos-QSVC [17]. Besides accuracy and F1-score, we record the complete confusion matrix and compute kernel Gram matrices from the first 150 training samples. These matrices are analyzed through their eigenvalue spectrum, condition number, and effective rank, defined as the minimum number of eigenvalues required to explain 99% of the total spectral energy. A run is classified as a recall collapse when precision minus recall exceeds 0.25, and as a trivial-positive classifier when recall is at least 0.999 while precision falls below 0.85. Both criteria are required because a precision–recall threshold alone identifies only four of the eight observed failures.

## IV. Results

### A. Accuracy and Variance

Classical models win on both datasets, and the margin widens as the data gets harder. Table 2 reports accuracy across all 56 quantum runs and 32 classical runs. On NSL-KDD, XGBoost averages 98.94% against 91.80% for the quantum classifier, a gap of 7.14 points. On ToN-IoT the gap grows to 16.42 points, 99.4% against 82.98%. The classical support vector machine, using the same seven principal components as the quantum classifier, still beats it by 5.57 and 12.52 points respectively, which rules out feature compression as the explanation.

TABLE II
ACCURACY ACROSS ALL RUNS. SD IS THE STANDARD DEVIATION ACROSS RANDOM SEEDS. QSVC (IDEAL) IS THE EXACT, NOISELESS STATEVECTOR KERNEL; QSVC (ALL) POOLS EVERY BACKEND

| Dataset | Model | Acc. | SD | Worst | $n$ |
|---|---|---|---|---|---|
| NSL-KDD | XGBoost | 0.9894 | 0.0008 | 0.9886 | 8 |
| NSL-KDD | SVM | 0.9737 | 0.0017 | 0.9712 | 8 |
| NSL-KDD | QSVC (ideal) | 0.9379 | 0.0103 | 0.9283 | 4 |
| NSL-KDD | QSVC (all) | 0.9180 | 0.0500 | 0.7560 | 24 |
| ToN-IoT | XGBoost | 0.9941 | 0.0013 | 0.9915 | 8 |
| ToN-IoT | SVM | 0.9550 | 0.0033 | 0.9503 | 8 |
| ToN-IoT | QSVC (ideal) | 0.8224 | 0.0711 | 0.7063 | 8 |
| ToN-IoT | QSVC (all) | 0.8298 | 0.0654 | 0.6947 | 32 |

Variance separates the two families more sharply than the mean does. XGBoost varies by 0.08 accuracy points across seeds on NSL-KDD and 0.13 points on ToN-IoT. The quantum classifier varies by 5.00 and 6.54 points, 62 and 49 times larger. A single accuracy number, averaged the way this literature typically reports it, hides that difference entirely.

### B. Discrete Failure Events

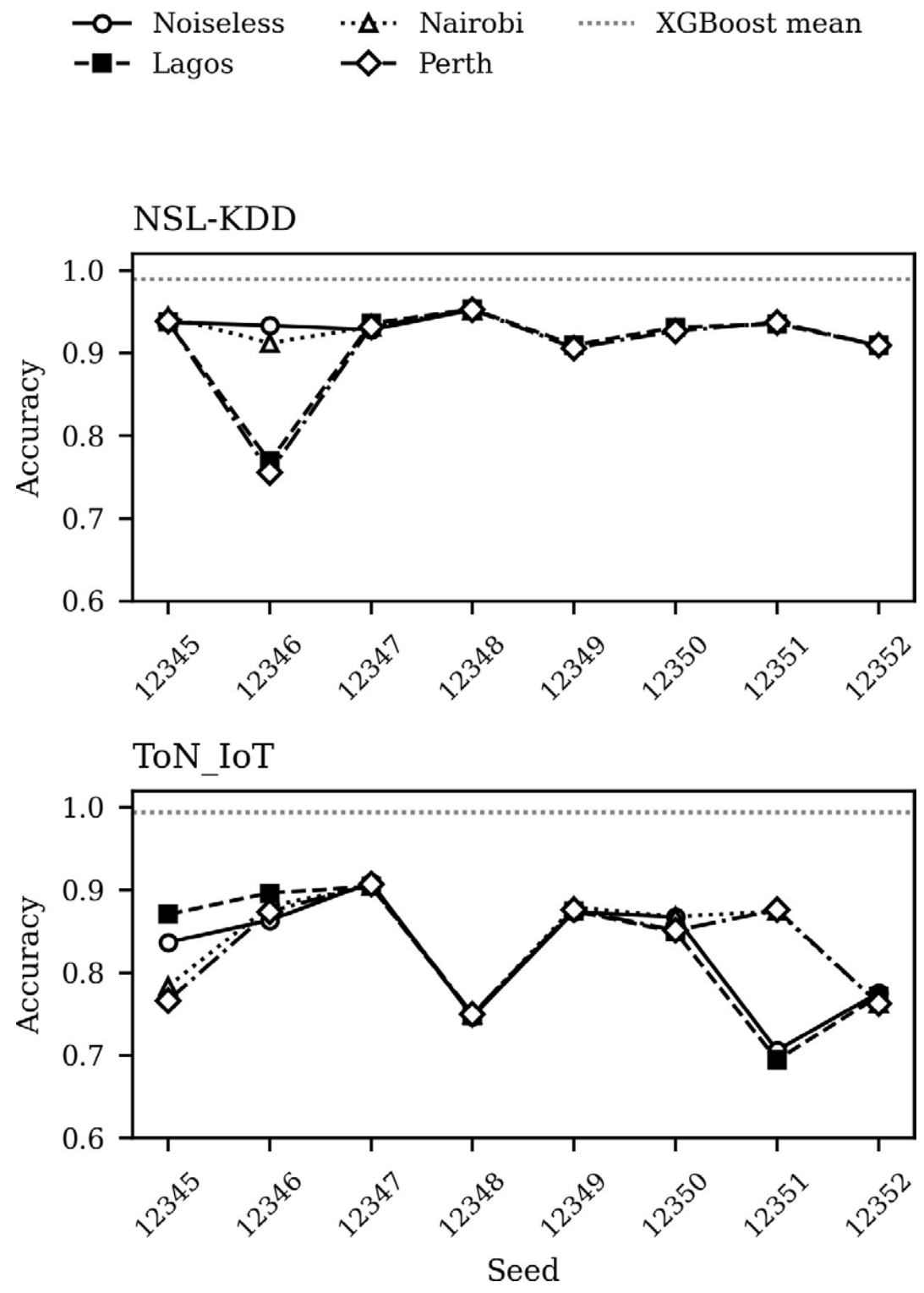

Fig. 2. Accuracy for every seed and backend. The dashed line marks the XGBoost mean. Both panels are flat until a specific seed drops sharply on some, but not all, backends.

Figure 2 makes the shape of the failure obvious before any table does. Accuracy is flat across seeds and backends until a specific seed hits, and then it falls sharply on some backends while its immediate neighbors stay healthy on the same seed. Eight of 56 runs stopped classifying this way, and Table 3 lists every one. Four are recall collapses, in which the model achieves perfect or near-perfect precision by refusing to flag anything it is not certain about. On NSL-KDD seed 12346 with the Perth noise model, the classifier missed 922 of 1,759 attacks, 52.4%, while raising zero false alarms. In a deployed sensor that is not conservatism. It is blindness.

The other four are the mirror image. On ToN-IoT seed 12352, every one of the four backends produced a classifier that labeled nearly all 6,332 test flows as attacks. Two of them, Nairobi and Perth, scored exactly 0.7631 accuracy, precisely the attack prevalence of the test split. A model that returns the majority class on every input reaches that number by construction, and it has learned nothing.

Noise is not the cause, and the noiseless backend proves it directly. ToN-IoT seed 12351 collapses on the noiseless kernel with recall 0.6173, and it also collapses on Lagos, but Nairobi and Perth remain healthy at recall above 0.91 on the identical seed. A noise explanation cannot produce that pattern, because the arm with zero noise fails while two noisier arms succeed on the same data split. Seed 12352 makes the point more plainly by degenerating on all four backends at once. Figure 3 places all 56 runs in precision–recall space, and the two failure modes appear as separate clusters pulling away from the diagonal in opposite directions.

TABLE III
ALL EIGHT DEGENERATE RUNS. TWO OF THE EIGHT, BOTH ON TON-IOT, OCCUR ON THE EXACT NOISELESS STATEVECTOR BACKEND. "RECALL" DENOTES A RECALL COLLAPSE; "TRIVIAL+" DENOTES A TRIVIAL-POSITIVE CLASSIFIER.

| Run | Seed | Recall | Prec. | Acc. | Mode |
|---|---|---|---|---|---|
| NSL/Lagos | 12346 | 0.504 | 1.000 | 0.769 | Recall |
| NSL/Perth | 12346 | 0.476 | 1.000 | 0.756 | Recall |
| ToN/Ideal | 12351 | 0.617 | 0.996 | 0.706 | Recall |
| ToN/Lagos | 12351 | 0.602 | 0.997 | 0.695 | Recall |
| ToN/Ideal | 12352 | 1.000 | 0.773 | 0.776 | Trivial+ |
| ToN/Lagos | 12352 | 1.000 | 0.770 | 0.772 | Trivial+ |
| ToN/Nairobi | 12352 | 1.000 | 0.763 | 0.763 | Trivial+ |
| ToN/Perth | 12352 | 1.000 | 0.763 | 0.763 | Trivial+ |

### C. Kernel Geometry

The kernel Gram matrix explains the mechanism behind both failure modes. Figure 4 plots effective rank by backend for both datasets. The noiseless kernel retains an effective rank of 46.5 out of 150 samples on NSL-KDD and 29.5 on ToN-IoT, meaning 99% of the spectral energy of a 150 by 150 Gram matrix sits in fewer than a third of its dimensions before any hardware is involved.

Its condition number reaches $4.84 \times 10^{12}$ on NSL-KDD and $3.00 \times 10^{13}$ on ToN-IoT. This is the exponential concentration signature that theory predicts for fidelity kernels [9], [10], observed here on real network traffic rather than a synthetic benchmark.

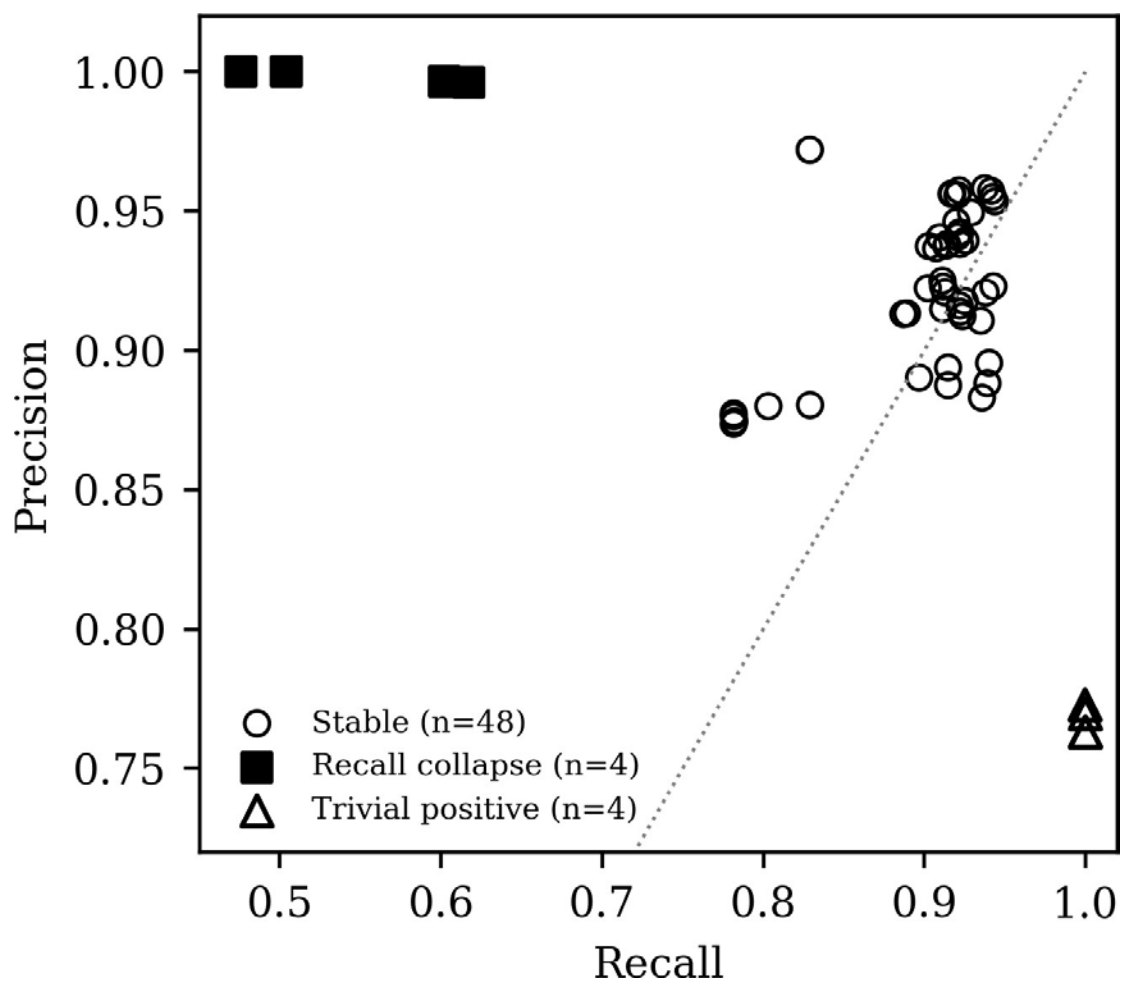

Fig. 3. Precision against recall for all 56 runs. Filled squares are recall collapses, open triangles are trivial-positive classifiers, and open circles are stable runs.

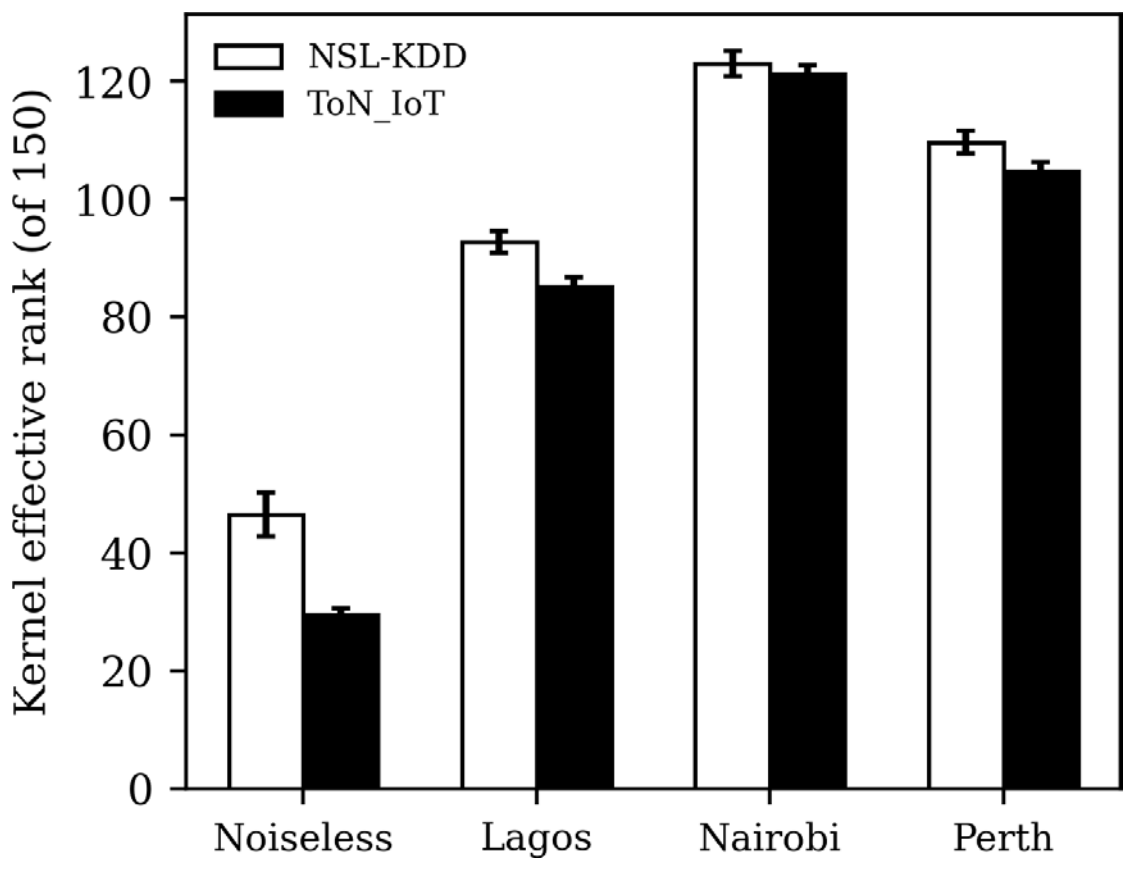

Fig. 4. Kernel effective rank, out of 150 samples, by backend and dataset. The noiseless kernel is the most concentrated in both datasets; every noisy backend raises effective rank well above it.

Noise changes this geometry, but not in the direction the prevailing narrative assumes. Adding a noise model raises effective rank from 46.5 to between 92.8 and 123.0 on NSL-KDD, and from 29.5 to between 85.3 and 121.2 on ToN-IoT, while condition number falls by eight to nine orders of magnitude. Decoherence pushes the Gram matrix toward the identity, which flattens the spectrum and superficially resembles a better-conditioned kernel. It is not a better kernel. It is a less informative one, and the accuracy numbers in Table 2 confirm that the added rank carries no signal.

One relationship does not hold, and we report it plainly rather than let it pass unnoticed. Effective rank explains the difference between backends and between datasets, but it does not predict which seed collapses within a backend. The correlation between per-run effective rank and per-run accuracy across all 56 runs is 0.01, indistinguishable from zero, and the two NSL-KDD collapse events sit at or above their backend average rank rather than below it. The concentration story explains why the kernel is fragile in aggregate. It does not explain why one specific seed breaks while its neighbor does not, and we leave that question open.

### D. Computational Cost

The cost side is not close, and it compounds the accuracy gap rather than offsetting it. A single noisy quantum run averages 1,472.7 seconds on NSL-KDD and 2,168.8 seconds on ToN-IoT. XGBoost trains and evaluates in 0.375 and 0.452 seconds on the same splits. That is a factor of 3,927 and 4,798 for a model that is 7.14 and 16.42 accuracy points worse and 49 to 62 times less stable across seeds. Even the noiseless quantum arm, which requires no sampling, takes 216.3 and 361.0 seconds on average.

## V. Discussion

The dataset with the more concentrated kernel is the dataset with the worse accuracy, and that relationship points at the encoding rather than the device. ToN-IoT carries 35 feature columns into the same seven qubits that NSL-KDD fills with 41, its noiseless effective rank is 29.5 against 46.5, and it contributes six of the eight degenerate runs. Compressing a wider and more categorical feature space through a fixed-width ZFeatureMap produces a flatter kernel, and a flatter kernel produces a less stable classifier. This is the behavior the concentration bounds predict [9], [10], and it scales with problem complexity rather than with device error rate.

Accuracy hides both failure modes, and that has a direct operational consequence for anyone deploying this class of model. Every degenerate run in Table 3 reports accuracy between 0.695 and 0.776, a range that reads as mediocre rather than broken. Only the confusion matrix reveals that one class was abandoned entirely. Any evaluation of quantum classifiers for security that reports headline accuracy without per-class recall will therefore describe a working system where none exists.

We propose a minimal reporting standard that follows directly from this evidence rather than from principle alone. Report worst-run accuracy alongside the mean, because Table 2 shows the mean can sit 20 points above the worst seed. Report seed-level variance, because a standard deviation 49 to 62 times larger than a classical baseline is itself a finding, not noise to be averaged away. Apply a two-sided degeneracy test, since a one-directional precision-minus-recall threshold missed four of our eight failures in Section 4.2. None of these three additions requires new experiments. They require reporting what the seed sweep already produced.

A fidelity kernel also complicates the audit trail a security team needs after a high-stakes decision, since neither classical SHAP-style tooling nor its quantum equivalents are yet mature enough to explain a single fidelity measurement the way a decision tree explains a split, a gap the field is only beginning to address [12].

For practitioners, the near-term conclusion is unambiguous on both datasets and at this qubit count. Gradient boosting is more accurate, thousands of times faster, and one to two orders of magnitude more stable across seeds. The open research question is not whether better hardware closes this gap. It is whether kernels engineered with inductive bias, such as projected or bandwidth-tuned constructions, escape the concentration that our measurements show is already present at seven qubits [9], [18].

## VI. Limitations

Our seed counts are small, four to eight runs per dataset and backend, so the 14.3% degeneracy rate carries wide error bars and should be read as evidence that the failure mode exists, not as a calibrated frequency. We use simulated noise models from calibration snapshots rather than live quantum hardware, so real device behavior may differ in ways a snapshot cannot capture. We evaluate one quantum method, a Pegasos classifier with a fidelity kernel and a ZFeatureMap, and our conclusions do not automatically transfer to projected kernels, covariant kernels, or variational classifiers built to resist concentration. The null relationship we report between effective rank and per-seed collapse holds within our own data, and a larger seed sweep is needed before ruling out a weaker or nonlinear version of that relationship. Finally, the noiseless arm uses an exact statevector kernel while the noisy arms use sampled fidelities, so the noiseless comparison isolates device error but not finite sampling.

## VII. Conclusion

This study evaluated Pegasos-QSVC for network intrusion detection through 56 runs on the NSL-KDD and ToN-IoT datasets, using one exact noiseless statevector backend and three noise-simulated superconducting backends. Under matched preprocessing, feature representation, data partitions, and evaluation conditions, Pegasos-QSVC underperformed XGBoost by 7.14 percentage points on NSL-KDD and 16.42 percentage points on ToN-IoT. It also exhibited substantially greater variation across random seeds.

More importantly, eight quantum runs degenerated into near-constant classifiers. Four experienced recall collapse, while four classified almost all observations as malicious. Two of these failures occurred on the noiseless backend, demonstrating that hardware noise is not a necessary condition for classifier degeneration. Analysis of the kernel Gram matrices further revealed low effective rank and severe ill-conditioning in the noiseless setting, indicating that representation-level spectral concentration is present before device noise is introduced. However, effective rank did not predict which individual seed would fail, leaving the causes of seed-specific collapse unresolved.

These findings show that mean accuracy alone is insufficient for evaluating quantum classifiers in security-critical applications. Future studies should report seed-level variability, worst-run performance, class-specific precision and recall, and explicit degeneracy tests. Further work should examine larger seed sets, live quantum hardware, and alternative kernel constructions with stronger inductive bias to determine whether the observed fragility can be reduced.